\documentclass[runningheads]{llncs}
\usepackage[mobile]{eccv}
\usepackage{eccvabbrv}

\usepackage{graphicx}
\usepackage{multirow}
\usepackage{booktabs}
\usepackage{colortbl}
\usepackage{xcolor}
\usepackage{float}
\PassOptionsToPackage{table}{xcolor}

\usepackage[accsupp]{axessibility}  

\usepackage[pagebackref,breaklinks,colorlinks,citecolor=eccvblue]{hyperref}

\usepackage{orcidlink}
\definecolor{darkgreen}{rgb}{0.0, 0.5, 0.0} 
\definecolor{darkred}{rgb}{0.6, 0.0, 0.0} 

\makeatletter
\DeclareRobustCommand\onedot{\futurelet\@let@token\@onedot}
\def\blfootnote{\xdef\@thefnmark{}\@footnotetext}
\makeatother

\begin{document}

\title{TrackDiffusion: Tracklet-Conditioned Video Generation via Diffusion Models}

\titlerunning{TrackDiffusion: Tracklet-Conditioned Video Generation via Diffusion Models}

\author{Pengxiang Li\inst{1}$^*$ \and
Kai Chen\inst{2}$^*$ \and
Zhili Liu\inst{2,4}$^*$ \and
Ruiyuan Gao\inst{3} \and
Lanqing Hong\inst{4} \and
\\
Guo Zhou\inst{5} \and
Hua Yao\inst{5} \and
Dit-Yan Yeung\inst{2} \and
Huchuan Lu\inst{1} \and
Xu Jia\inst{1}$^\dagger$}

\authorrunning{P.~Li et al.}

\institute{Dalian University of Technology \and
Hong Kong University of Science and Technology \and
The Chinese University of Hong Kong \and
Huawei Noah's Ark Lab \quad $^{5}$\ GAC Research and Development Center \\
\small\url{https://kaichen1998.github.io/projects/trackdiffusion/}
}

\maketitle


\vspace{-6mm}
\begin{abstract}
Despite remarkable achievements in video synthesis, achieving granular control over complex dynamics, such as nuanced movement among multiple interacting objects, still presents a significant hurdle for dynamic world modeling, compounded by the necessity to manage \textit{appearance and disappearance}, \textit{drastic scale changes}, and ensure \textit{consistency for instances across frames}.
These challenges hinder the development of video generation that can faithfully mimic real-world complexity, limiting utility for applications requiring high-level realism and controllability, including advanced scene simulation and training of perception systems.
To address that, we propose \textbf{TrackDiffusion}, a novel video generation framework affording fine-grained trajectory-conditioned motion control via diffusion models, which facilitates the precise manipulation of the object trajectories and interactions, overcoming the prevalent limitation of scale and continuity disruptions. 
A pivotal component of TrackDiffusion is the \textit{instance enhancer}, which explicitly ensures inter-frame consistency of multiple objects, a critical factor overlooked in the current literature.
Moreover, we demonstrate that generated video sequences by our TrackDiffusion can be used as training data for visual perception models.
To the best of our knowledge, this is the first work to apply video diffusion models with tracklet conditions and demonstrate that generated frames can be beneficial for improving the performance of object trackers.

\vspace{-2mm}
\end{abstract}

\blfootnote{
$^*$ Equal contribution. Contact: \email{lipengxiang@mail.dlut.edu.cn} \\
$^\dagger$ Corresponding author.
}


\vspace{-10mm}

\section{Introduction}
\label{sec:intro}
\begin{figure}[t]
    \centering
    \includegraphics[width=\linewidth]{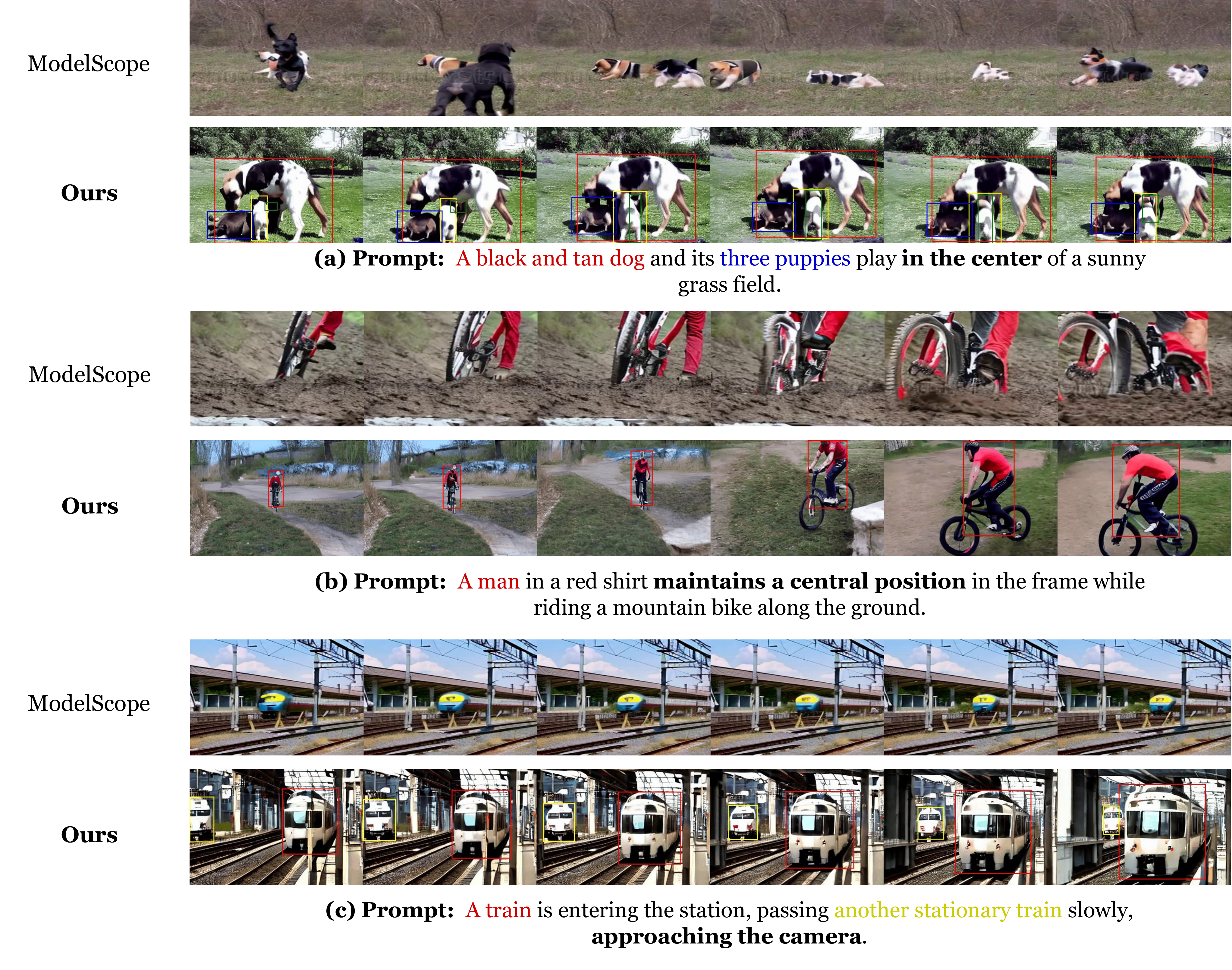}
    \vspace{-8mm} 
    \caption{\textbf{Qualitative comparison on the trajectory-conditioned motion control}. 
    ModelScope~\cite{Wang_Yuan_Chen_Zhang_Wang_Zhang} does not support controls other than text. In contrast, the generation results of \textit{TrackDiffusion} are more consistent with the input prompts.}
    \label{fig:teaser}
    \vspace{-6mm} 
\end{figure}

Benefiting from the development of the diffusion models, video generation has achieved breakthroughs, particularly in text-to-video (T2V) generation models~\cite{wu2022nuwa, hong2022cogvideo}. The utilization of diffusion models and large-scale text-video pairs markedly expanded the ability to generate diverse video content~\cite{luo2023videofusion,blattmann2023align,chen2024videocrafter2,zhang2023show,kondratyuk2023videopoet,girdhar2023emu} enabling a more nuanced interpretation of textual prompts and translating them into dynamic, visually compelling narratives.
Although textual descriptions provide a friendly interactive manner for image generation, it is not easy for them to impose fine control over generated content. Several control signals have been employed to generate images with more flexibility and higher quality,  such as control signals from segmentation, content edges~\cite{zhang2023adding}, and object boxes~\cite{li2023gligen, chen2023integrating, gao2023magicdrive}. to specify object or image layout. Considering video's nature of continuity and temporal dynamics, textual descriptions also can not provide sufficient information to guarantee highly realistic details, even Sora~\cite{openai2023video} may fail in case of spontaneous appearances of objects~\cite{bloomberg_video}. Thus, fine-grained motion control could contribute much to high-quality video generation.
Such fine-grained control not only enhances visual quality but also has the potential to enable applications like perception model training, animated storytelling, and advanced user interfaces.

While fine-grained motion control is a natural interaction for video generation, it is still under-explored, especially for diffusion-based video generation models.
Despite the progress~\cite{wang2023motionctrl, yin2023dragnuwa} in the field, existing generative models often fail to maintain instance-level consistency across frames critical for reproducing the complex temporal dynamics found in natural settings.
Consequently, they struggle to capture the dynamic interplay among multiple objects, especially in complex scenarios marked by occlusion, overlapping objects, and unpredictable rapid movements as depicted in Fig.~\ref{fig:teaser}.

In this work, we introduce \textit{TrackDiffusion}, a novel framework specifically designed to fill this lacuna.
Integrating with video diffusion models, \textit{TrackDiffusion} enables fine-grained motion control of generated contents with object boxes.
Specifically, we first introduce instance-aware location tokens for each object, which embed identity information of boxes into boxes across frames, and are helpful in addressing the object occlusion and re-occurrence.
Besides, one distinctive component of our framework is the \textit{instance enhancer} module. This simple yet effective component provides inter-frame consistency of objects, ensuring remarkable instance-level consistency.
Finally, gated cross-attention is employed to seamlessly integrate the box conditions into a pretrained video diffusion model such that the huge amount of computation for training from scratch could be avoided.

Our extensive experiments demonstrate that \textit{TrackDiffusion} surpasses prior methods in the quality of the generated video data.
Furthermore, ablation studies confirm the necessity of introducing instance-aware location tokens and instance enhancer for achieving these results. Our experiment also shows that generated videos by \textit{TrackDiffusion}, as augmented data, could benefit tracking tasks and bring further improvement on the performance of tracking models.

The main contributions of this work contain three parts:
\begin{enumerate}
  \item We present the \textbf{very first known} application of DMs to generate continuous video sequences directly from the tracklets, a methodological innovation that transcends the capabilities of existing video generative models.
  \item A novel component of our framework, the \textit{instance enhancer}, is proposed to provide consistent inter-frame object identity, even in challenging conditions such as occlusion and rapid movement.
  \item Our experimental results demonstrate that by incorporating tracklet constraints, the quality of the videos improves substantially, and the track average precision (TrackAP) score of the object tracker, which assesses the alignment between the given boxes and the generated objects, experiences a significant boost, underscoring the efficacy of motion control.
\end{enumerate}

\section{Related Work}
\label{sec:relatedwork}
\subsubsection{Layout-to-Image Generation} (L2I), focusing on converting high-level graphical layouts into photorealistic images, has witnessed considerable advancements. 
GLIGEN~\cite{li2023gligen} enhances the pre-trained diffusion models with gated self-attention layers for improved layout control, while LayoutDiffuse~\cite{cheng2023layoutdiffuse} employs novel layout attention modules tailored for bounding boxes. 
Instead, GeoDiffusion~\cite{chen2023integrating} enables various geometric controls directly via text prompts to support object detection \cite{han2021soda10m,li2022coda} data generation, 
which is further extended for concept removal by Gemo-Erasing~\cite{liu2023geomerasing}, and 3D geometric control by MagicDrive~\cite{gao2023magicdrive}.

\vspace{-2mm}
\subsubsection{Text-to-Video Generation}
(T2V), following the successful trajectory of the text-to-image (T2I) generation, has achieved significant advancements. 
Most of the T2V methodologies~\cite{he2023latent,ho2022imagen,zhou2022magicvideo} tend to focus on depicting the continuous or repetitive actions from textual prompts, rather than capturing the dynamics of multiple, changing actions or events. However, these methods generally lack the ability to generate complex transitions and diverse event sequences. On the other hand, recent works such as LVD~\cite{lian2023llm} employ large language models to create dynamic scene layouts for video diffusion, concentrating on text-driven layout generation. VideoComposer~\cite{wang2024videocomposer} enables conditions such as sketches, depth maps, and motion vectors.
VideoDirectorGPT~\cite{Lin_Zala_Cho_Bansal_2023} and DriveDreamer~\cite{Wang_Zhu_Huang_Chen_Lu} have furthered multi-scene video generation, showcasing advancements in the field. 
Different from previous works, our \textit{TrackDiffusion} presents the very first attempt to explicitly encode trajectory conditions into the video generation process, revealing superior flexibility for user controls and interaction.

\section{Method}
\label{sec:method}

\begin{figure*}[t]
    \centering
    \includegraphics[width=\textwidth]{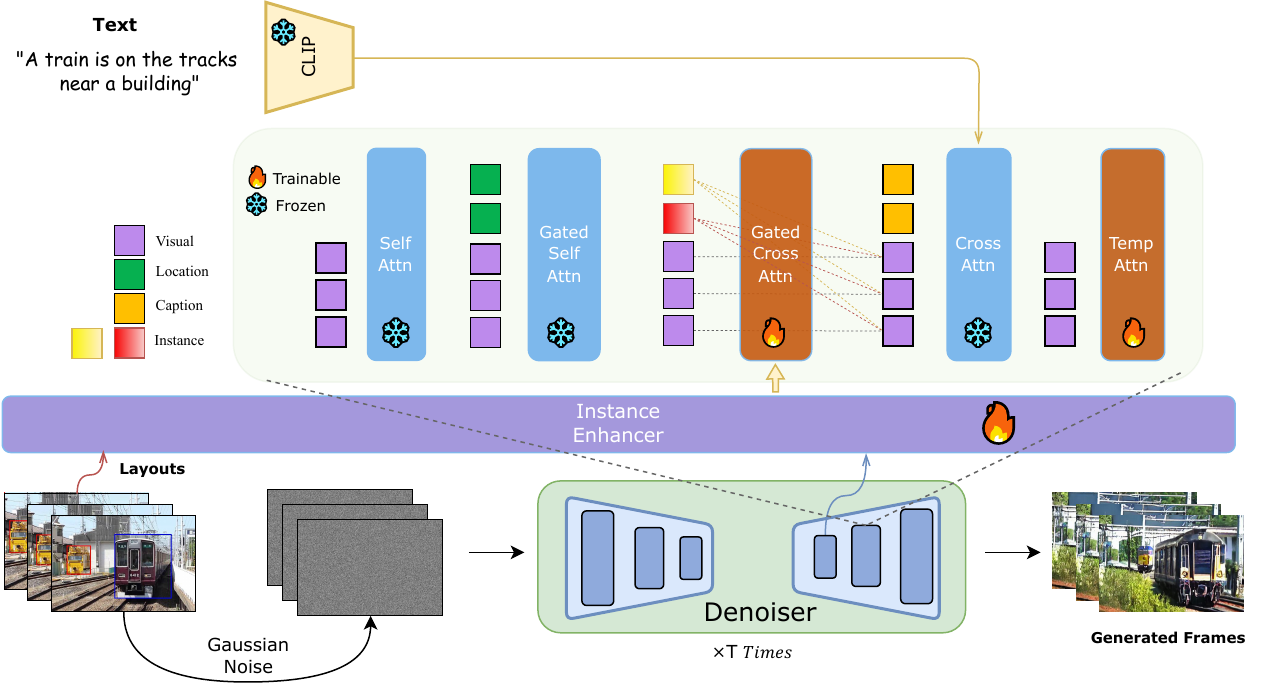}
    \vspace{-6mm} 
    \caption{\textbf{Model architecture of \textit{TrackDiffusion}}.
    The framework generates video frames based on the provided tracklets and employs the \textit{Instance Enhancer} to reinforce the temporal consistency of foreground instance. A new gated cross-attention layer is inserted to take in the new instance information.
    }
    \vspace{-2mm}
    \label{fig:pipeline}
\end{figure*}


In this section, we first introduce the latent diffusion model (LDM~\cite{rombach2021highresolution}), on which our method is based, in Section~\ref{sec:preliminary}. Then, we introduce the \textit{TrackDiffusion} pipeline, the \textit{instance-aware location tokens} and \textit{temporal instance enhancer}, in Section~\ref{sec:condition}. We also present our methods for enhancing instance consistency across frames, particularly in the video clips with noticeable spatial changes. 
An overview of \textit{TrackDiffusion} is shown in Fig.~\ref{fig:pipeline}.


\subsection{Preliminary: Latent Diffusion Models (LDM)}\label{sec:preliminary}

Recent advancements in image synthesis have been significantly driven by LDM. These models excel by focusing on the distribution within the latent space of images, marking a notable leap in performance in this domain. The LDM comprises two main components: an autoencoder and a diffusion model. 

The autoencoder is responsible for compressing and reconstructing images, utilizing an encoder \( \mathcal{E} \) and a decoder \( \mathcal{D} \). Specifically, the encoder projects an image \( x \) into a lower-dimensional latent space \( z \), followed by the decoder reconstructing the original image from this latent representation. The reconstruction process yields an image \( \tilde{x} = \mathcal{D}(z) \), approximating the original image \( x \).
Given that the data distribution \(z_0 \sim q\left(z_0\right)\) is progressively corrupted by Gaussian noise over \( T \) steps, this process follows a variance schedule denoted by \( \beta_1, \ldots, \beta_T \):


\begin{equation}
\begin{aligned}
q\left(z_t \mid z_{t-1}\right)=\mathcal{N}\left(z_t ; \sqrt{1-\beta_t} z_{t-1}, \beta_t \mathbb{I}\right), \quad t=1, \ldots, T
\end{aligned}
\end{equation}
with a U-Net, $\epsilon_\theta\left(z_t; t \right)$, trained to predict this added noiseusing a loss function:
\begin{equation}
L(\theta)=\mathbb{E}_{t \sim \mathcal{U}(1, T), \epsilon_t \sim \mathcal{N}(0,1)}\left[\left\|\epsilon_t-\epsilon_\theta\left(z_t ; t, \mathbf{y}\right)\right\|^2\right],
\end{equation}
where $x_t$ is the noisy sample of $x_0$ at timestep $t$. The condition $\mathbf{y}$ can be $\varnothing$ (unconditional generation), text~\cite{rombach2021highresolution} or images~\cite{ho2022cascaded}, etc.

\subsection{Tracklet-Conditioned Video Generation}\label{sec:condition}

\subsubsection{Overview.}
Our method, \textit{TrackDiffusion}, introduces an innovative approach to video generation from tracklets, addressing the challenges of instance consistency and spatial-temporal coherence in complex video sequences. At its core, \textit{TrackDiffusion} leverages the strengths of LDM while introducing novel mechanisms to ensure fidelity in instance-level consistency across video frames. The methodological backbone of \textit{TrackDiffusion} consists of four pivotal components: \textit{Instance-Aware Location Tokens}, \textit{Temporal Instance Enhancer}, \textit{Motion Extractor}, and \textit{Gated Cross-Attention}.
Together, these components form a synergistic framework that not only captures the intricacies of individual frames but also preserves the natural flow and continuity of multi-object interactions across a video sequence. By effectively addressing the challenges of spatial and temporal coherence, our \textit{TrackDiffusion} sets a new standard in the domain of the tracklet-conditioned video generation, simultaneously offering a solution for training and generating consistent video sequences with complicated object movement.

\vspace{-2mm}
\subsubsection{Task Definition.}
The primary objective of \textit{TrackDiffusion} is to generate high-fidelity video sequences from tracklets, where a tracklet refers to a sequence of object bounding boxes across frames, coupled with their respective category information. Formally, given a set of tracklets $\mathcal{T} = \{\tau_1, \tau_2, \ldots, \tau_n\}$, where each tracklet $\tau_i$ corresponds to an object instance across $T$ frames, our task is to generate a video sequence $V$ that accurately represents the motion and appearance of these instances.
Each tracklet $\tau_i$ is defined as $\tau_i = \{(b_{i,1}, c_{i,1}), (b_{i,2}, c_{i,2}), \ldots, (b_{i,T}, c_{i,T}) \}$, where $b_{i,t}$ denotes the bounding box coordinates of the $i$-th instance in frame $t$, and $c_{i,t}$ represents the category of the instance. The generated video $V$ is a sequence of frames $\{v_1, v_2, \ldots, v_T\}$, where each frame $v_t$ is a synthesis of the instances as per their tracklet descriptions at time $t$.

\vspace{-2mm}
\subsubsection{Instance-Aware Location Tokens.}
To fully leverage the condition of object layouts, the coordinates of a 2D object box \( b_{i,t} \) in a frame are projected into the embedding space similarly to the positional encoding in GLIGEN~\cite{li2023gligen}. This projection, \( B_{i,t} = \text{Fourier}(b_{i,t}) \), is then concatenated with the box's category embedding and transformed into the conditioning representation \( H_{i,t} \),
as:

\begin{equation}
    H_{i,t} = \text{MLP}(\left[ c_{i,t}, B_{i,t} \right]),
\end{equation}
where $c_{i,t}$ denotes the category embedding of the $i$-th bounding box computed with the CLIP model~\cite{radford2021learning}, and $\text{Fourier}(\cdot)$ refers to Fourier embedding~\cite{Mildenhall_Srinivasan_Tancik_Barron_Ramamoorthi_Ng_2020}. 

To encourage consistency of instances across frames, we further complement the layout condition representation with instance identity information. In this way, the trajectories of various instances in the sequence could be distinguished, and the continuity of instances across frames would be reinforced. The enhanced instance-aware location token is represented as \( H'_{i,t} = H_{i,t} + e_i \), with \( e_i \) represents a learnable token for the instance denoted by the \( i \)-th box in the frame.

Subsequently, a gated self-attention~\cite{li2023gligen} is employed to impose conditional layout information into visual features as shown below:
\begin{equation}
    V_t = V_t + \tanh(\beta) \cdot \text{TS}(\text{SelfAttn}(\left[ V_t, H'_{:,t} \right])),
\end{equation}
where \( V_t \) denotes the visual feature tokens in frame \( t \), \( H'_{:,t} \) represents the enhanced location tokens for all boxes at time \( t \), \(\beta\) is a trainable parameter, and \(\text{TS}(\cdot)\) is the token selection operation focusing only on visual tokens at each time frame.
We refer readers to check~\cite{li2023gligen} for more details.

\begin{figure}[t]
    \centering
    \includegraphics[width=0.8\linewidth]{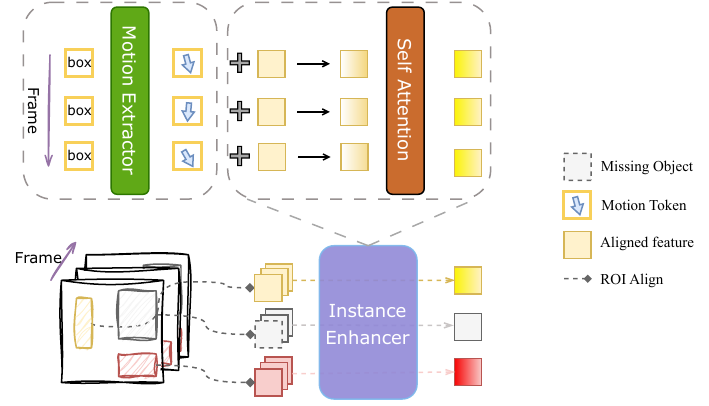}
    \vspace{-3mm}
    \caption{\textbf{Illustration of Temporal Instance Enhancer}. Each instance's latent features, after \textit{ROI Align}, are concatenated with motion tokens and then processed through self-attention layers. We demonstrate the specific enhance operation using the yellow instance as an example.
    }
    \label{fig:Enhancer}
    \vspace{2mm}
\end{figure}

\subsubsection{Temporal Instance Enhancer.}
In the context of tracklet conditioned video generation, a significant challenge is to maintain consistency of generated instances across frames, especially when instances have large spatial displacement in the sequence. 
Existing works employ temporal attention to encourage temporal consistency over frames where attention works at each position along time. However, when there is dramatic motion with the object or the camera, the attention computed on the same position would not work very well (see Fig.~\ref{fig:videocomp}).
To address this challenge, we propose a novel temporal attention called \textit{Temporal Instance Enhancer} where attention is computed on the same instance instead of position, as explained in the following.

Let \( F_i \) represent the multi-dimensional feature tensor for the $i$-th instance, constructed by temporally concatenating features extracted from each frame \( t \) using the provided bounding box through ROIAlign. This process aligns the features of the same instance across all temporal frames to the same spatial size:

\begin{align}
F_i = \bigoplus_{t=1}^{T} \text{ROIAlign}(V_t, B_{t,i}),
\end{align}
where \( \bigoplus \) denotes the concatenation operation, and \( T \) is the number of frames.
In addition, the representation for each multi-dimensional feature tensor is further added with a trajectory motion representation $P_i$, as defined in the following. We simultaneously use a box consistent with the latent shape to perform ROIAlign, representing the background feature tensor.
\subsubsection{Motion Extractor.}
The trajectory representation is computed by applying self-attention to a sequence of box embeddings, each representing the same instance across successive frames. This method not only captures the spatial coordinates of the instance at each time step (\ie, the location information) but also, by analyzing these sequences, discerns the movement of the instance over time (\ie, the motion information). The inclusion of self-attention enables the model to effectively track objects smoothly through occlusions and interactions by inferring the continuity of object presence and movement. It is the model's ability to detect subtle shifts in position and temporal dependencies through self-attention that facilitates the accurate modeling of motion trajectories, as it comprehends the dynamic changes in an instance's location from frame to frame.

Similar to temporal attention, self-attention is then applied to the feature representation for each instance along time, as shown in Eqn.~\ref{eq:tracklet_rep},~\ref{eq:tmp_inst_attn} and Fig.~\ref{fig:Enhancer}.
\begin{equation}
    P_i = \text{SelfAttn}({B_{1,i}, B_{2,i}, \ldots, B_{T,i}})
    \label{eq:tracklet_rep}
\end{equation}
\begin{equation}
    F'_i = \text{SelfAttn}(F_i \oplus P_i),
    \label{eq:tmp_inst_attn}
\end{equation}
where \( F'_i \) represents the enhanced features for $i$-th instance cube, which plays an important role in promoting accurate and consistent video generation.

\vspace{-2mm}
\subsubsection{Gated Cross-Attention.}

To integrate the enhanced instance features in video generation, we borrow the idea from GLIGEN and design a gated cross-attention layer which is inserted after the gated self-attention layer. This layer can make full use of temporally enhanced instance features for consistent video generation.
The visual feature tokens from the gated self-attention layer are represented as \( V = [v_1, \ldots, v_M] \), where \( M \) denotes the total number of tokens in the flattened latent. Then the gated cross-attention layer could be simply formulated as below:
\begin{equation}
    V = V + \text{CrossAttn}([V, F'_i]), \label{eqn:gated_cross}
\end{equation}
where \( F'_i \) denotes the enhanced instance features. This insertion ensures that the visual tokens in the LDM framework are now additionally informed by the aggregated instance features, thereby maintaining the consistency of instance appearance across different frames.

\section{Experiments}

\begin{table*}[t]
    \small
    \centering
    \begin{tabular}{@{} lc ccccc @{}}
        \toprule
        \multirow{2}{*}{\textbf{Method}} & \multirow{2}{*}{\textbf{Output}} & \multicolumn{3}{c}{\textbf{YoutubeVIS}} \\
        \cmidrule(lr){3-5}
        &  \textbf{Res.} & \textbf{FVD↓} & \textbf{TrackAP↑} & \textbf{TrackAP$_{50}$↑}\\
        \midrule
        Oracle$^{*}$ & - & - & 45.4 & 64.1\\
        \hline
        CogVideo(Eng.)~\cite{hong2022cogvideo} & 160$\times$160 & 1384 & - & -\\
        LVDM~\cite{he2023latent} & 256$\times$256 & 1011 & - & -\\
        ModelScopeT2V~\cite{Wang_Yuan_Chen_Zhang_Wang_Zhang} & 256$\times$256 & 786 & - & - \\
        ZeroScopeT2V~\cite{Wang_Yuan_Chen_Zhang_Wang_Zhang} & 576$\times$320 & 750 & - & -\\
        Show-1~\cite{zhang2023show} & 576$\times$320 & 704 & - & -\\
        VideoCrafter~\cite{chen2023videocrafter1} & 256$\times$256 & 690 & - & -\\
        VideoComposer~\cite{wang2024videocomposer} & 256$\times$256 & 738 & 19.8 & 30.6\\
        \hline
        \textbf{Vanilla} & 256$\times$256 & 603 & 36.0 & 56.2\\
        \textbf{TrackDiffusion} & 256$\times$256 & 605 & 39.4\textcolor{darkred}{(+3.4)} & 62.0\textcolor{darkred}{(+5.8)}\\
        \rowcolor{gray!20} \textbf{TrackDiffusion} & 480$\times$320 & \textbf{548} & \textbf{44.7}\textcolor{darkred}{(+8.7)} & \textbf{68.0}\textcolor{darkred}{(+11.8)}\\
        \bottomrule
    \end{tabular}
    \vspace{2mm}
    \caption{\textbf{Comparison of generation fidelity on YoutubeVIS datasets.} 
    Vanilla is our customized baseline, similar to DriveDreamer~\cite{Wang_Zhu_Huang_Chen_Lu}, fine-tuned on the YTVIS2021 dataset. *: represents the real image \textit{Oracle} baseline.}
    \label{tab:comparison_main}
\end{table*}

Given that the traditional text-to-video (T2V) datasets (\eg, MSR-VTT~\cite{xu2016msr} and UCF101~\cite{soomro2012ucf101}) typically do not contain bounding box annotations, we turn to adopt tracking datasets that offer precise tracklet annotations. This allows us to appraise our model's performance in generating text-to-video content within the multi-object tracking scenarios, where tracklet-conditioned video synthesis is essential. Our primary quantitative evaluations are conducted using a version of our model that extends the ModelScopeT2V~\cite{Wang_Yuan_Chen_Zhang_Wang_Zhang} framework.

\begin{figure*}[t]
    \centering
    \includegraphics[width=1.0\textwidth]{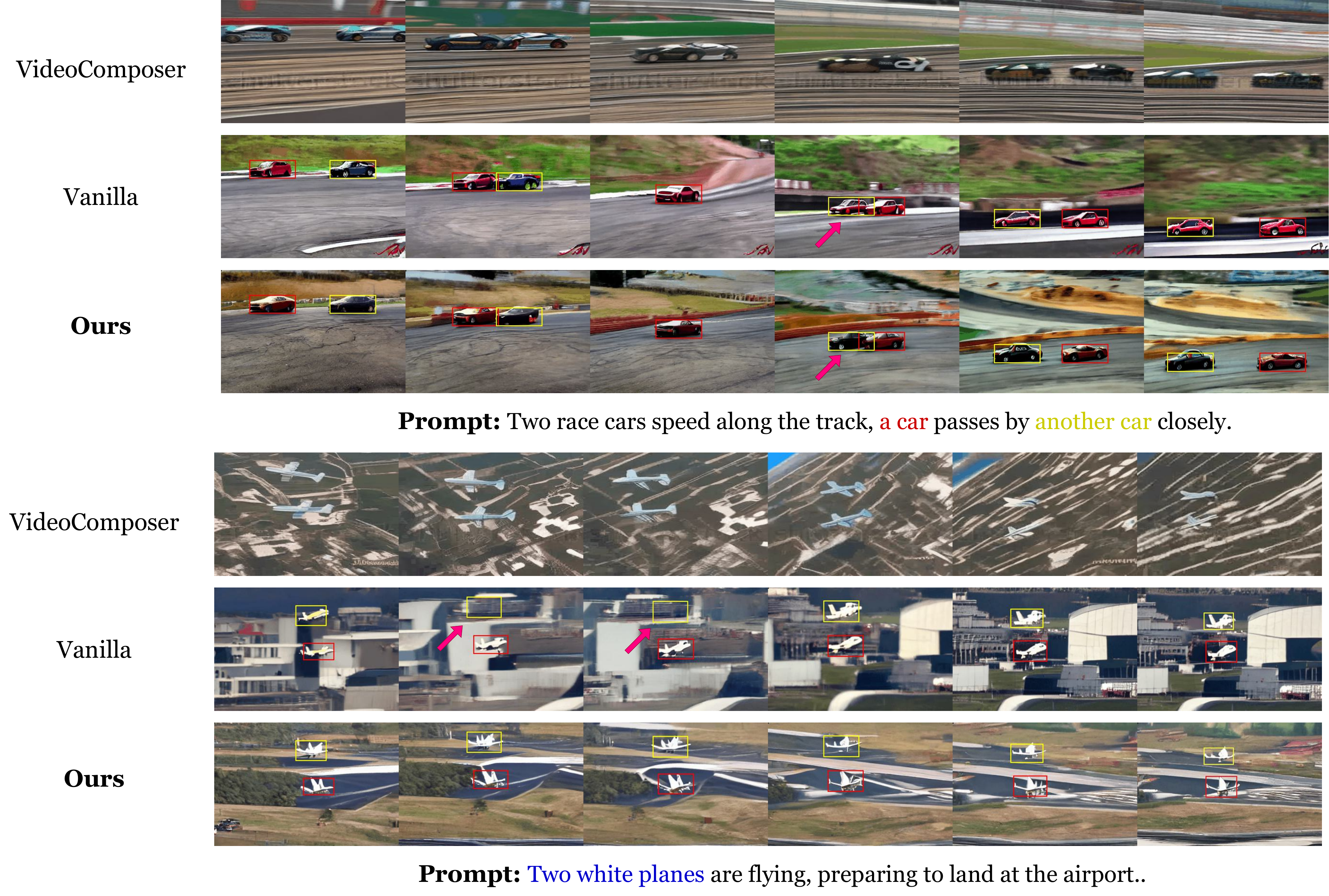}
    \caption{
    \textbf{Qualitative comparison} among different methods. The input text prompt is shown on the bottom side of the figure.}
    \vspace{-4mm}
    \label{fig:videocomp}
\end{figure*}

\subsection{Implementation Details}\label{sec:implement}
\subsubsection{Dataset.}
Our experiments purposefully utilize both the YTVIS2021~\cite{vis2021} and the MOT-17~\cite{milan2016mot16} datasets with a unified objective: to validate effectiveness of our approach in maintaining the consistency across multi-object generation. The YTVIS2021 dataset, serving as a cornerstone in VIS literature, includes 2,985 training videos with high-quality bounding box annotations from 40 semantical classes. On the other hand, the MOT-17 is a prominent dataset in MOT research, encompassing over 10,000 frames that focus on the pedestrian tracking. For the YTVIS2021 dataset, due to the absence of annotations for the validation set, we have randomly selected 160 videos from the training set to serve as our validation set. Regarding MOT-17, we divide the MOT-17 training dataset in half, using one half for training and the other half for validation, following common practice. To compensate the absence of captions in the YTVIS2021 and MOT-17 datasets, we utilized the VideoBLIP~\cite{li2023blip} model to generate captions for each video clip.

\vspace{-2mm}
\subsubsection{Evaluation metrics.}
In our evaluation, we utilize the captions and box annotations from the validation sets of YTVIS. We adopt \textit{FVD}~\cite{unterthiner2018towards} to evaluate video quality. To evaluate the grounding accuracy, which is the correspondence between the input bounding box and the generated instance, we use the \textit{Tracking Average Precision} (\textit{TrackAP}~\cite{Yang_Fan_Xu_2019}). This involves using pretrained QDTrack~\cite{Fischer_Pang_Huang_Qiu_Chen_Darrell_Yu_2022} model to track objects in the generated videos, which are then compadarkred with the ground truth tracklets. It's important to note that since previous text-to-video methods do not support incorporating the box annotations as input, it is not equitable to compare them on this metric. Therefore, we limit our comparison to report FVD scores for reference.
FVD scores on more datasets (\eg, UCF101) are provided in Appendix~\ref{app:exp}.

\vspace{-2mm}
\subsubsection{Baseline.}
We introduce a baseline model, \textbf{Vanilla}, in Tab.~\ref{tab:comparison_main}, conceptually motivated by the DriveDreamer~\cite{Wang_Zhu_Huang_Chen_Lu} framework. This model represents our specialized adaptation for layout-conditioned video generation, incorporating key principles from the DriveDreamer methodology, yet distinctly engineered to align with the realm of layout-conditioned video generation on YoutubeVIS.

\subsubsection{Details.}
We implement our approach based on the Diffusers~\cite{von-platen-etal-2022-diffusers} code base for ModelScopeT2V. Our training methodology comprises two stages. In \textbf{Stage 1}, we focus on \textit{single-image layout controllability} by removing all temporal layers and employing gated self-attention. \textbf{Stage 2} \textit{extends the approach to video data}, introducing \textit{temporal attention}, \textit{temporal convolution}, and an \textit{instance enhancer} for video-level control. We trained the \textit{Stage 1} on the corresponding training set for 60,000 steps, the \textit{Stage 2} for 50,000 steps. The training process was carried out on 8 NVIDIA Tesla 80G-A800 GPUs. During generation, frames are sampled using the DPM-Solver~\cite{Lu_Zhou_Bao_Chen_Li_Zhu_2022} scheduler for 50 steps with the classifier-free guidance (CFG) set as 5.0.

\subsection{Comparison with Existing T2V Methods}
\begin{figure*}[t]
    \centering
    \includegraphics[width=1.0\textwidth]{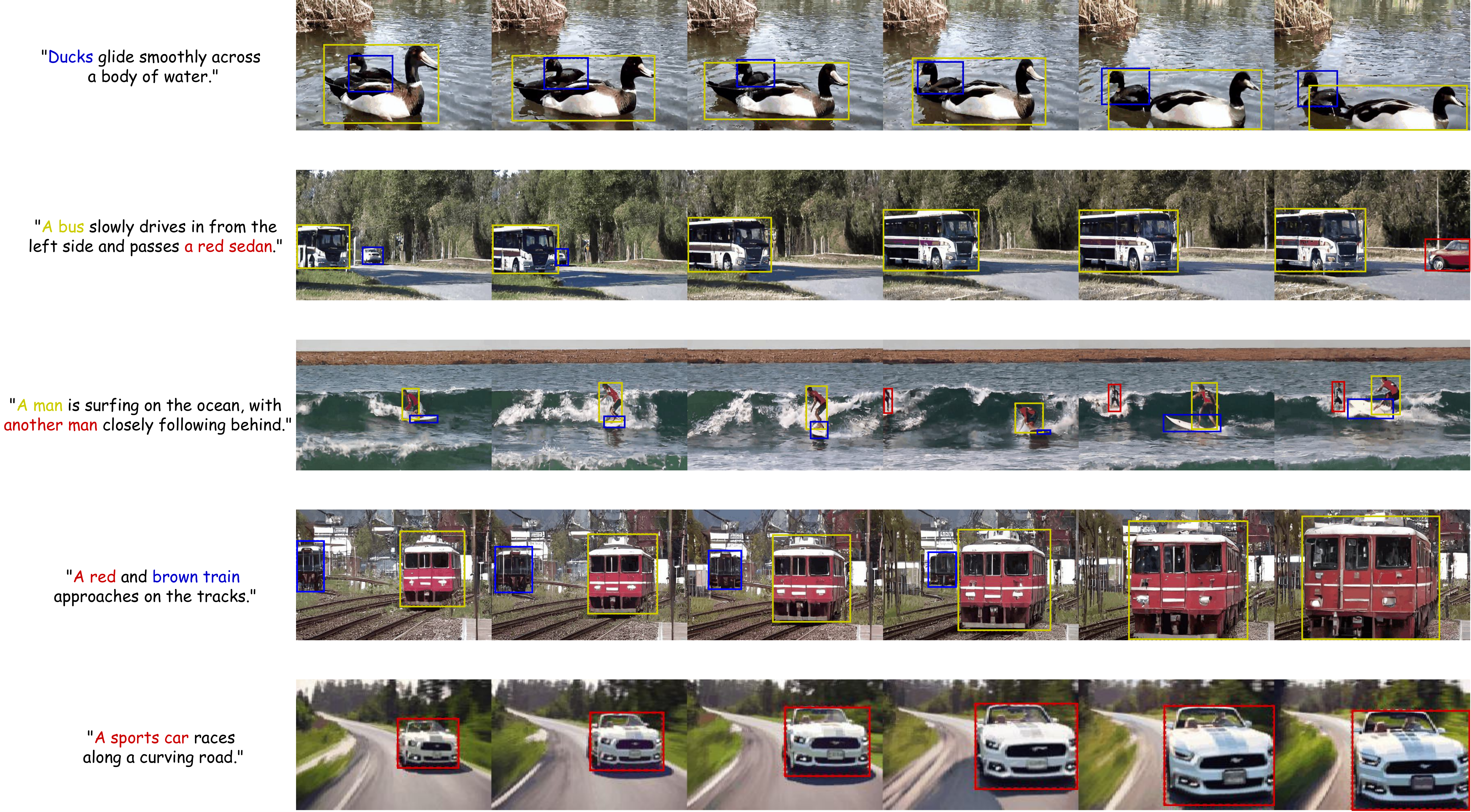}
    \caption{\textbf{Video generation with tracklet conditions on YTVIS2021}. The input text prompt is shown on the left side of the figure. Different instances' bounding boxes are distinguished by different colors.}
    \vspace{-2mm}
    \label{fig:YTVIS_demo}
\end{figure*}

In this section, we conduct a comprehensive evaluation of the proposed \textit{TrackDiffusion} with regard to video quality and trajectory controllability, which demands a realistic representation of objects while being consistent with geometric layouts. Our visual results are shown in Fig.~\ref{fig:YTVIS_demo}.

\begin{figure*}[t]
    \centering
    \includegraphics[width=1.0\textwidth]{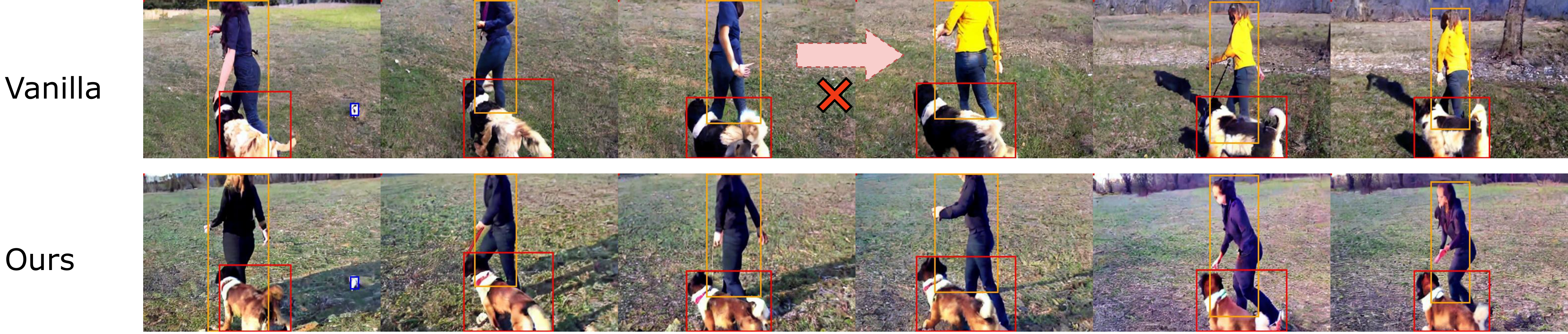}
    \caption{
    \textbf{Severe appearence changes.}
    In the entire video sequence, the Vanilla method is prone to changes in appearance, while \textit{TrackDiffusion} achieves better consistency in the appearance of instances.}
    \label{fig:hard_case_2}
\end{figure*}

\subsubsection{Video Quality.}
\label{section:Quality}
Tab.~\ref{tab:comparison_main} compares our \textit{TrackDiffusion} model with a range of recent video synthesis methods on the YTVIS dataset. Our approach demonstrates a significant advancement in the field, evidenced by competitive FVD scores which serve as a metric for video quality. Specifically, \textit{TrackDiffusion} at a resolution of 256$\times$256 achieves an FVD score of 605, showcasing its effectiveness in synthesizing high-fidelity videos. Notably, an enhanced version of TrackDiffusion, operating at a higher resolution of 480$\times$320, further improves the FVD score to 548.
This performance surpasses several contemporary models, including Vanilla, VideoCrafter, and others, underscoring the significance of the fine-grained control and consistency mechanisms implemented in \textit{TrackDiffusion}. Notably, while the standard version aligns with the resolution of many counterparts, the high-resolution variant sets a new benchmark in the field. This is primarily because, at higher resolutions, the instance enhancer can extract cleaner instance features, laying a solid foundation for improved video quality.
These results underscore the capability of \textit{TrackDiffusion} to not only maintain but also enhance the quality of video synthesis, even when scaling to higher resolutions. This indicates that the model can effectively handle increased complexity and detail, a critical factor for realistic video generation.

Furthermore, as demonstrated in Fig.~\ref{fig:videocomp}, we present the generation results of VideoComposer, Vanilla, and \textit{TrackDiffusion}. For VideoComposer, we control video generation using depth maps. 
Vanilla struggles to maintain consistency in the appearance of instances.
The color of the racing car changes in the first example, and in the second example, the object is even lost. Despite the control signals provided by the depth maps, VideoComposer's granularity remains coarse. Due to the absence of box-level control, the quality of the generated objects is somewhat inferior, and it fails to accurately produce high-quality videos that align with the user intended motion control.

\subsubsection{Trajectory Controllability.}
Tab.~\ref{tab:comparison_main} also showcases the trajectory control precision of various video synthesis models, with a particular focus on the TrackAP metric from the YoutubeVIS dataset. \textit{TrackDiffusion}, in both its standard and high-resolution variants, demonstrates a superior ability to precisely control trajectory. The standard 256$\times$256 resolution version of \textit{TrackDiffusion} achieves a TrackAP score of 39.4, which is an improvement of 3.4 points over the Vanilla model. When the resolution is increased to 480$\times$320, \textit{TrackDiffusion}'s performance further improves, reaching a TrackAP score of 44.7, which marks an 8.7 point increase.
It's important to note that TrackAP scores in our experiments do not equate to the success rate in trajectory control. For comparison, we tested tracker performance using real data, as indicated by the Oracle results, which we consider an approximate upper bound for TrackAP scores. The closer our TrackAP scores are to the Oracle benchmark, the better the generated data's fidelity. Therefore, we should concentrate on the comparative difference in TrackAP scores between methods rather than their absolute values.

\subsection{Ablation Study}
To ascertain the effectiveness of our proposed design moudles, we conducted an ablation study focusing on crucial components of the model, such as the instance tokens and the instance enhancer. These evaluations were carried out using the YTVIS validation set, as discussed in Sec.~\ref{sec:implement}.

\vspace{-3mm}
\subsubsection{Setup.}
We conduct ablation studies primarily focusing on fidelity and report the results for FVD and TrackAP metrics. To balance training duration and the adverse effects of lower resolutions on the tracker, our experiments in this section generate videos at a resolution of \(384 \times 256\). We employ Mask2Former~\cite{cheng2021mask2former} to evaluate TrackAP, mitigating the impact of data noise.
Check more ablation results in Appendix~\ref{app:ablation}.

\begin{table}[t]
\centering
\setlength{\tabcolsep}{0.7mm}{
\begin{tabular}{@{}lccccc@{}}
\toprule
Setting & InstEmb & T-Enhancer & FVD↓  & TrackAP↑ & TrackAP$_{50}$↑ \\ 
\midrule
(a)     & & & 741  & 34.2    & 60.4      \\
(b)     & \(\checkmark\) & &  765  & 35.0\textcolor{darkred}{(+0.8)} & 61.7\textcolor{darkred}{(+1.3)}      \\
\rowcolor{gray!20}
(c)& \(\checkmark\) & \(\checkmark\) &  \textbf{729}  & \textbf{38.7}\textcolor{darkred}{(+4.5)} & \textbf{64.3}\textcolor{darkred}{(+3.9)}      \\
\bottomrule
\end{tabular}
}
\vspace{2mm}
\caption{\textbf{Ablation of the instance embedding and the temporal instance enhancer.}
Defaults are marked in \colorbox{gray!20}{gray}.
}
\vspace{-2mm}
\label{tab:ablation_study}
\end{table}

\begin{figure}[t]
    \centering
    \includegraphics[width=0.9\linewidth]{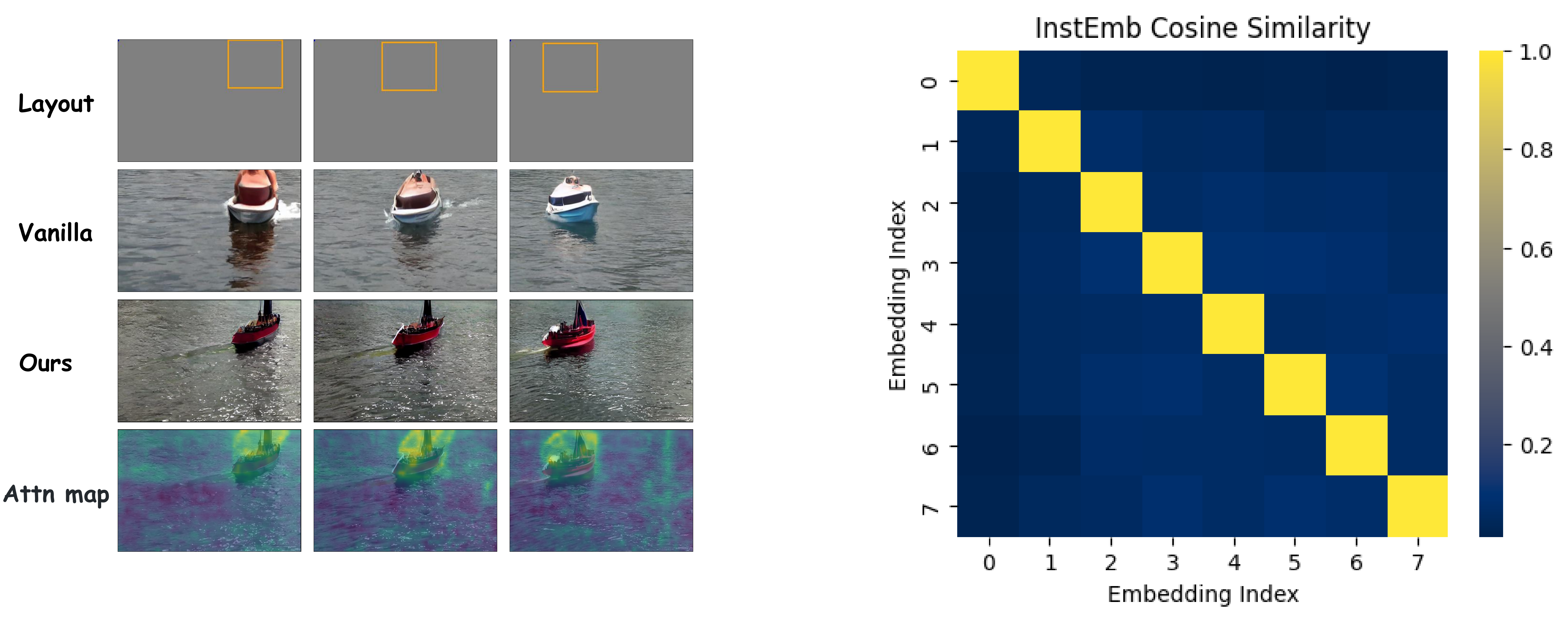}
    \caption{
    \textbf{Instance Embedding and Temporal Consistency Visualization.}
    When there are significant changes in object locations, \textit{TrackDiffusion} can better maintain the instances temporal consistency.
}
    \label{fig:attn_vis}
\end{figure}

\subsubsection{Effectiveness of Consistency Module.}

The ablation study in Tab.~\ref{tab:ablation_study} examines the impact of incorporating instance embeddings on instance consistency, using FVD and TrackAP as metrics. In Setting (a), without instance embeddings, the model shows baseline performance. However, introducing instance embeddings in Setting (b) leads to a slight increase in FVD (from 741 to 765), suggesting a minor trade-off in video quality. Despite this, TrackAP improves by 0.8, indicating enhanced instance consistency. This trade-off may be attributed to the additional parameters and not yet fully optimized training. Notably, when provided with sufficient training time, as shown in Table~\ref{tab:comparison_main}, instance embedding does not negatively impact FVD. The most significant improvements are seen in Setting (c), where both instance embeddings and temporal instance enhancer are employed, further confirming effectiveness of these features in improving instance consistency (see more details in Fig.~\ref{fig:hard_case_2}).

In Fig.~\ref{fig:attn_vis} (left), the ``Attn map'' row demonstrates the superimposition of the original image with the attention map, and we can observe that the highlighted regions in the attention map overlap with the target areas in the image. This alignment serves as evidence that the features extracted are being correctly attended to. Moreover, the instance embedding cosine similarity matrix confirms our model's capability to differentiate between distinct instances, preventing incorrect associations in sequence generation. Low similarity scores outside the main diagonal reflect this effective disambiguation.

\subsubsection{Effectiveness of Motion Extractor.}
We manually curate a subset of 23 videos from the validation set, ensuring that each video encompasses instances of target overlap or re-occurrence to varying degrees. We aim to demonstrate the effectiveness of the proposed motion information extraction through experiments on this subset, using FVD and TrackAP as metrics.
Results indicate a marked improvement when the motion extractor is employed, as evidenced by lower FVD and higher TrackAP scores. Specifically, the inclusion of the motion extractor yields a decrease in FVD to \textbf{774} from 793 and an increase in TrackAP by \textbf{2.0} points, achieving a score of 36.5. This enhancement is also reflected in the TrackAP\(_{50}\) score, which sees an increase of \textbf{2.5} points, reaching 59.0. These results corroborate the efficacy of the motion extractor in our model, signifying its essential role in capturing and maintaining coherent motion trajectories in complex video scenes.

\begin{table}[tbp]
  \centering
  \begin{tabular}{l|cccc}
    \toprule
    \multirow{2}{*}{\textbf{Method}} & \multicolumn{2}{c}{\textbf{YoutubeVIS}} & \multicolumn{2}{c}{\textbf{MOT-17}} \\
    \cmidrule(lr){2-3} \cmidrule(lr){4-5}
    & \textbf{TrackAP} & \textbf{TrackAP$_{50}$} & \textbf{MOTA} & \textbf{HOTA} \\
    \midrule
    Real only & 45.4 & 64.1 & 67.0 & 56.3\\
    \midrule
    Vanilla & 45.2\textcolor{darkgreen}{(-0.2)} & 60.7\textcolor{darkgreen}{(-3.4)} & 65.8\textcolor{darkgreen}{(-1.2)} & 54.3\textcolor{darkgreen}{(-2.0)}\\
    TrackDiffusion &  46.7\textcolor{darkred}{(+1.3)} & 65.6\textcolor{darkred}{(+1.5)} & 68.4\textcolor{darkred}{(+1.4)} & 57.5\textcolor{darkred}{(+1.2)}\\
    \bottomrule
  \end{tabular}
  \vspace{2mm}
  \caption{\textbf{Comparison of trainability on YTVIS and MOT-17 datasets}. The 480$\times$320 \textit{TrackDiffusion} variant is used for trainability evaluation.}
  \label{tab:real_data}
\end{table}
\subsection{Synthetic data for data augmentation}

In this section, we explore the advantages of using frames generated by \textit{TrackDiffusion} for training object trackers. We incorporate these generated frames as augmented data in the training process of an object tracker, aiming to assess the effectiveness of our proposed model more thoroughly.

\subsubsection{Setup.}
Utilizing the data annotations from the YTVIS or MOT-17 training set as input, we generate frames that serve as augmented samples. These are then merged with real images to enrich the training dataset. Subsequently, a QDTrack, initialized with LVIS~\cite{gupta2019lvis} pre-trained weights, is trained following the standard 1× schedule. The performance of this model is then evaluated on the validation set. To increase generality, we also organized similar experiments on MOT-17 dataset.
Experiments on more trackers are provided in Appendix~\ref{app:exp}.

\vspace{-2mm}
\subsubsection{Discussion.}
The results in Tab.~\ref{tab:real_data} demonstrate the effectiveness of \textit{TrackDiffusion} in enhancing trainability for object tracking models. When comparing the models trained with only real data to those augmented with \textit{TrackDiffusion} generated frames, a noticeable improvement is observed. Specifically, \textit{TrackDiffusion} increases TrackAP by 1.3 and TrackAP$_{50}$ by 1.5, indicating a significant enhancement in tracking accuracy. In contrast, the Vanilla method shows a slight decrease in performance, with TrackAP and TrackAP$_{50}$ dropping by 0.2 and 3.4, respectively.


\section{Conclusion}
In conclusion, our work presents \textit{TrackDiffusion}, a novel approach to generating continuous video sequences from tracklets, effectively utilizing diffusion models for video synthesis in the context of multi-object tracking. Our model introduces innovative mechanisms, including Instance-Aware Location Tokens and Temporal Instance Enhancer, which together facilitate the generation of high-quality and temporally consistent video sequences. The experimental results also show the potential of \textit{TrackDiffusion} in enhancing the training of perception models, thereby marking a significant step forward in the realm of synthetic video data generation. Future work will focus on addressing the outlined limitations, further improving the model's generalization capabilities, and exploring its applicability in a broader range of real-world scenarios.

\subsubsection{Acknowledgement}
We gratefully acknowledge support of MindSpore, CANN (Compute Architecture for Neural Networks) and Ascend AI Processor used for this research.

\bibliographystyle{splncs04}
\bibliography{main}
\clearpage
\clearpage
\appendix

\setcounter{table}{0}
\setcounter{figure}{0}
\renewcommand{\thetable}{A\arabic{table}}
\renewcommand{\thefigure}{A\arabic{figure}}

\section*{Supplementary Material}
\section{More Experiments}
\label{app:exp}
\subsection{Fidelity}
\subsubsection{UCF-101.}
We follow the experiment settings of PixelDance~\cite{zeng2023make} to evaluate on UCF-101. Specifically, we sampled 2,048 videos from the UCF-101 test set, generating descriptive text prompts for each of them. To obtain the tracklet of objects, we use Grounding DINO~\cite{liu2023grounding} to get bounding boxes on the first frame and MixFormer~\cite{cui2022mixformer} to extend these boxes to subsequent frames.
Tab.~\ref{tab:ucf} presents our results on the UCF-101 dataset, which are consistent with the trends observed on the YTVIS dataset. \textit{TrackDiffusion} achieves better FVD and TrackAP scores compared to other methods. However, the absolute values of the AP scores on UCF-101 are generally lower. This could be attributed to the fact that our automatic annotation approach may not be sufficiently refined and lacks manual refinement, resulting in lower quality annotations.

\begin{table*}[ht]
    \small
    \centering
    \begin{tabular}{@{} lcccc @{}}
        \toprule
        \multirow{2}{*}{\textbf{Method}}  & \multicolumn{3}{c}{\textbf{UCF-101}} \\
        \cmidrule(lr){2-4}
        & \textbf{FVD↓} & \textbf{TrackAP↑} & \textbf{TrackAP$_{50}$↑}\\
        \midrule
        MagicVideo~\cite{zhou2022magicvideo} & 699 & - & - \\
        LVDM~\cite{he2023latent}  & 641 & - & -\\
        Make-A-Video~\cite{singer2022make} & 367 & - & -\\
        VidRD~\cite{gu2023reuse} & 363 & - & -\\
        PYOCO~\cite{ge2023preserve} & 355 & - & -\\
        \midrule
        ModelScopeT2V~\cite{Wang_Yuan_Chen_Zhang_Wang_Zhang} & 410 & 0.7 & 1.2 \\
        VideoComposer~\cite{wang2024videocomposer} & 452 & 8.2 & 13.4 \\
        \hline
        \textbf{Vanilla} & 360 & 16.2 & 28.7 \\
        \rowcolor{gray!20} \textbf{TrackDiffusion} & \textbf{293} & \textbf{20.5} & \textbf{32.6}\\
        \bottomrule
    \end{tabular}
    \vspace{2mm}
    \caption{UCF-101 zero-shot text-to-video generation. ModelScopeT2V does not support box input, and the values in the table are only used as a reference for comparison.}
    \label{tab:ucf}
\end{table*}
\vspace{-8mm}

\subsubsection{GOT10K.}
To further validate the robustness of our approach and mitigate the potential impact of noisy annotations in the UCF-101 dataset, we extend our experiments to the GOT10K dataset~\cite{Huang2021}. GOT10K is a widely-recognized benchmark in the Single Object Tracking (SOT) community, consisting of over 10,000 video sequences capturing diverse moving objects in real-world scenarios. We used the GOT10k validation set, which contains 180 videos, and then generated 900 16-frames videos.
\begin{table*}[ht]
    \small
    \centering
    \begin{tabular}{@{} lcccc @{}}
        \toprule
        \multirow{2}{*}{\textbf{Method}}  & \multicolumn{3}{c}{\textbf{GOT10K}} \\
        \cmidrule(lr){2-4}
        & \textbf{FVD↓} & \textbf{Success↑} & \textbf{Normalized Precision↑}\\
        \midrule
        Oracle$^{*}$ & - & 84.4 & 96.0\\
        \hline
        ModelScopeT2V~\cite{Wang_Yuan_Chen_Zhang_Wang_Zhang} & 635 & - & - \\
        ZeroScopeT2V~\cite{Wang_Yuan_Chen_Zhang_Wang_Zhang}  & 641 & - & -\\
        VideoCrafter~\cite{chen2023videocrafter1} & 603 & - & -\\
        VideoComposer~\cite{wang2024videocomposer} & 630 & 55.2 & 62.1\\
        \hline
        \textbf{Vanilla} & 627 & 75.8 & 87.6 \\
        \rowcolor{gray!20} \textbf{TrackDiffusion} & \textbf{532} & \textbf{79.4} & \textbf{93.2}\\
        \bottomrule
    \end{tabular}
    \caption{\textbf{Comparison of generation fidelity on GOT10K datasets.} Vanilla is our custom baseline, similar to DriveDreamer. *: represents the real \textit{Oracle} baseline.}
    \label{tab:got10k_compare}
\end{table*}

\begin{table}[ht]
\centering
\begin{tabular}{@{}lccccc@{}}
\toprule
Method & Train Set & MOTA↑  & HOTA↑ & IDF1↑ & IDSw.↓ \\ 
\midrule
SORT~\cite{bewley2016simple} & M17ht & 62.0 & 52.0 & 57.8 & 5847 \\
\rowcolor{gray!20}
SORT & M17ht+TD & \textbf{65.1} & \textbf{53.3} & \textbf{58.5} & \textbf{5556} \\
\midrule
ByteTrack~\cite{zhang2022bytetrack} & M17ht & 75.2 & 65.2 & 76.9 & 672 \\
\rowcolor{gray!20}
ByteTrack & M17ht+TD & \textbf{76.8} & \textbf{65.8} & \textbf{77.9} & \textbf{597} \\
\midrule
StrongSORT++~\cite{du2023strongsort} & M17ht & 74.9 & 68.4 & 81.9 & 534 \\
\rowcolor{gray!20}
StrongSORT++ & M17ht+TD & \textbf{77.9} & \textbf{69.7} & \textbf{84.1} & \textbf{393} \\
\bottomrule
\end{tabular}
\caption{\textbf{Comparison of various MOT baselines.} M17ht refers to the ``mot17halftrain'' dataset, and TD refers to TrackDiffusion.}
\vspace{-3mm}
\label{tab:mot}
\end{table}

\subsection{Trainability}
\subsubsection{MOT-17.}
We further conduct trainability experiments on the MOT-17 dataset following the exact same setting with Sec.~4.4, but with more MOT baselines considered.
Following common practice, we divide the MOT-17 training dataset into one half for training and the other one for evaluation using the standard MOT metrics (\ie, MOTA, HOTA, IDF1, and IDSw.) based on mmtracking.
As reported in Tab.~\ref{tab:mot}, consistent improvement is observed with the help of video sequences generated by TrackDiffusion, revealing effectiveness of our method.

\subsection{Driving Video Generation.}
\subsubsection{nuScenes.}
In addition to our primary models, we extend our training to include real driving scene (RDS) data. The training data is sourced from the real-world driving dataset nuScenes~\cite{Caesar_Bankiti_Lang_Vora_Liong_Xu_Krishnan_Pan_Baldan_Beijbom_2020}, which consists of 700 training videos and 150 validation videos. Each video in the dataset captures approximately 20 seconds of footage using six surround-view cameras, providing a comprehensive view of the driving environment.

The nuScenes dataset provides annotations in the form of the 2Hz 3D bounding boxes. To utilize this data, we first convert the bounding boxes using scripts provided by MMDetection3D~\cite{mmdet3d2020}. Subsequently, we conduct our training based on the annotations derived from these keyframes. Due to resource limitations, our training exclusively employed data from the front camera. We employed the same settings as those used for training on the YTVIS dataset, with a training resolution of \(480 \times 320\).

As presented in Tab.~\ref{tab:nus}, we compared our proposed \textit{TrackDiffusion} method with baseline approaches. Our method achieved FVD scores of 364, significantly outperforming the baseline models. These results demonstrate considerable advantages over other approaches, illustrating the efficacy of \textit{TrackDiffusion} in capturing temporal consistency and generating high-quality video sequences.

\begin{figure}[t]
    \centering
    \includegraphics[width=\linewidth]{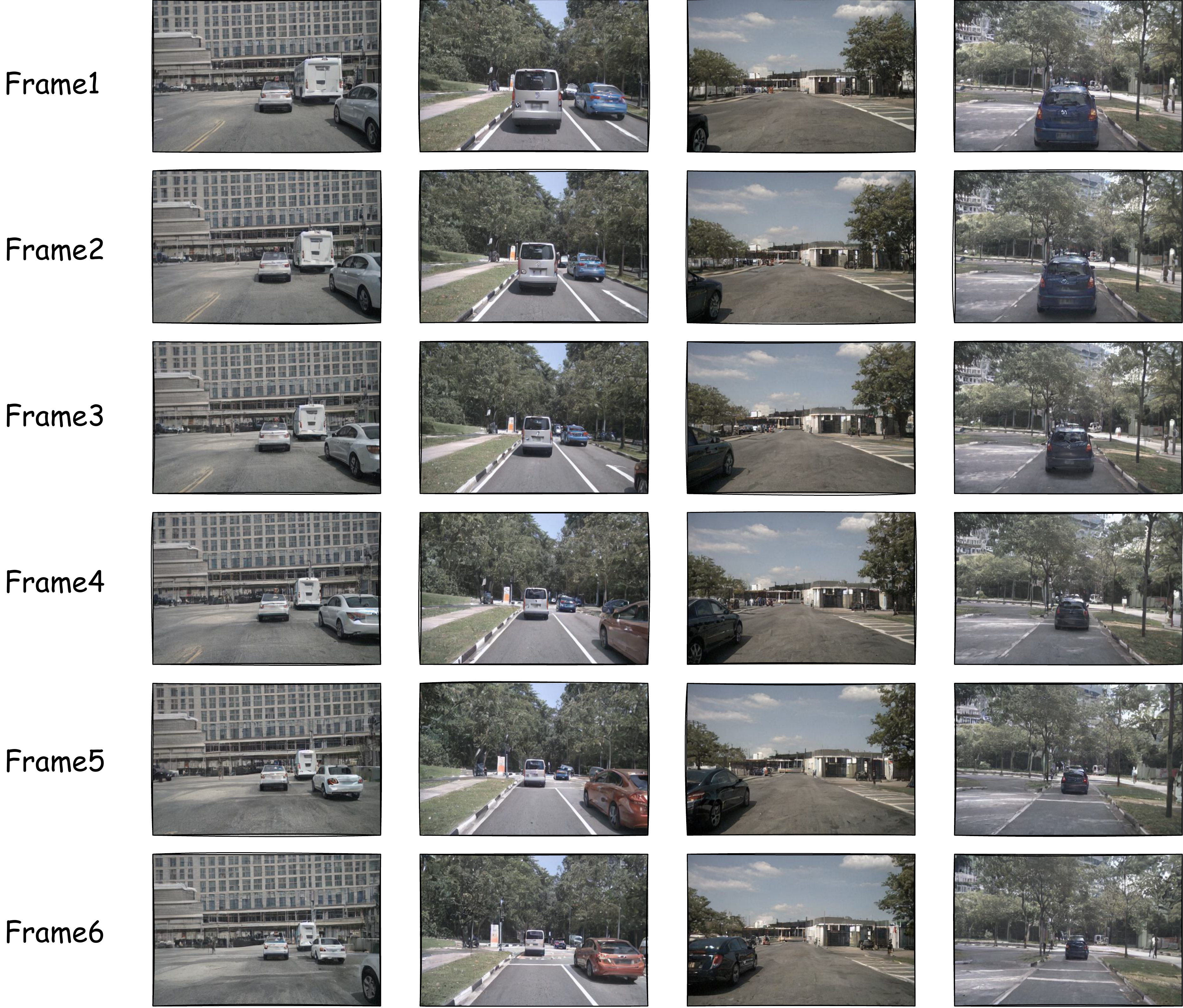}
    \caption{\textbf{More qualitative results on the nuScenes dataset.} Our method achieved commendable instance consistency even when trained solely on keyframes, demonstrating its effectiveness in maintaining the consistency of instance appearance.
    }
    \label{fig:nus_demo}
    \vspace{-4mm}
\end{figure}

\begin{table}[ht]
  \centering
  \begin{tabular}{l|cccc}
    \toprule
    Method & InstEmb & T-Enhancer & FVD\\
    \midrule
    \textbf{Vanilla} & & & 456 \\
    \rowcolor{gray!20}
    \textbf{TrackDiffusion} & \(\checkmark\) & \(\checkmark\) & \textbf{364}\\
    \bottomrule
  \end{tabular}
  \vspace{2mm}
  \caption{\textbf{Comparison on nuScenes validation set.}}
  \label{tab:nus}
\end{table}

\section{More Ablation Study}\label{app:ablation}

\subsubsection{Ablation on gated cross-attention.}
As shown in the main paper Fig.~2 and Eqn.~8, our approach utilizes gated cross-attention to absorb the instance features. We can also consider gated self-attention~\cite{li2023gligen}. This design variation was ablated on the YTVIS2021 dataset. As demonstrated in Tab.~\ref{tab:self_cross}, the gated self-attention results in a similar FVD of 760, but yields a lower TrackAP of 37.2, compared to 38.7 for the cross-attention. This shows that cleaner instance information, isolated from visual tokens, enables the model to utilize features more effectively, which can be attributed to the more focused and undiluted use of instance features in cross-attention, thereby enhancing the model's capability for maintaining instance consistency in videos.

\label{sec:extra_abl}

\begin{table}[t]
\centering
\begin{tabular}{@{}lccccc@{}}
\toprule
Setting & FVD↓ & TrackAP↑ & TrackAP$_{50}$↑ \\ 
\midrule
self-attn     & 765 & 37.2 & 63.9 \\
\rowcolor{gray!20}
cross-attn    & \textbf{760} & \textbf{38.7}\textcolor{darkred}{(+1.5)} & \textbf{64.3}\textcolor{darkred}{(+0.4)} \\
\bottomrule
\end{tabular}
\vspace{2mm}
\caption{\textbf{Ablations on gated cross-attention.}}
\label{tab:self_cross}
\end{table}
\begin{table}[b]
\centering
\begin{tabular}{@{}lccccc@{}}
\toprule
Setting & FVD↓ & TrackAP↑ & TrackAP$_{50}$↑ \\ 
\midrule
Encoder Block     & 816 & 38.2 & 63.6 \\
\rowcolor{gray!20}
Decoder Block    & \textbf{762} & \textbf{38.7}\textcolor{darkred}{(+0.5)} & \textbf{64.3}\textcolor{darkred}{(+0.7)} \\
\bottomrule
\end{tabular}
\vspace{2mm}
\caption{\textbf{Comparison for positions of the instance enhancer.}}
\label{tab:instance_enhancer_position}
\end{table}
\vspace{-5mm}

\subsubsection{Ablation on instance enhancer position.}
In this ablation study, we evaluate the impact of the instance enhancer's position within our \textit{TrackDiffusion} model. The study involves two main configurations: integrating the instance enhancer into the encoder blocks and into the decoder blocks of the U-Net architecture. Due to resource limitations, we did not attempt to include the instance enhancer in both the encoder and decoder simultaneously.

The comparative results indicate a notable improvement when the instance enhancer is positioned in the decoder block, as evidenced by the lower FVD and higher TrackAP. The FVD score decreases from 816 to 762, suggesting enhanced video quality and consistency. Moreover, improvements in TrackAP (from 38.2 to 38.7) and TrackAP$_{50}$ (from 63.6 to 64.3) imply better consistency.

This enhancement can be attributed to the decoder block's role in reconstructing and refining the output frames. By placing the instance enhancer in the decoder block, it seems to have a more significant impact on the visual attributes of instances, leading to improved model performance in terms of both fidelity and tracking precision. These findings suggest a crucial role of the decoder block in handling instance-specific details.

\subsubsection{Ablation on necessity of real data.}
We randomly sample 10\%, 50\%, and 75\% of the real training set, and each subset is utilized to train a QDTrack together with generated frames separately. The number of gradient steps are maintained unchanged for each pair experiment with the same amount of real data by enlarging the batch size adaptively.

As shown in Tab.~\ref{tab:necessity}, \textit{TrackDiffusion} achieves consistent improvement over different real training data budgets. The more scarce real data is, the more significant improvement \textit{TrackDiffusion} achieves, sufficiently revealing that generated frames do help ease data necessity. With only 75\% of real data, the tracker performs comparably with the full real dataset by augmenting with \textit{TrackDiffusion}.
\begin{table}[t]
  \centering
  \begin{tabular}{l|cccc}
    \toprule
    Method & 10\% & 50\% & 75\% & 100\% \\
    \midrule
    w/o TrackD & 32.7 & 42.1 & 43.1 & 45.4 \\
    \textbf{w/ TrackD} & 37.2\textcolor{darkred}{(+4.5)} & 43.4\textcolor{darkred}{(+1.3)} & 44.6\textcolor{darkred}{(+1.5)} & 46.7\textcolor{darkred}{(+1.3)}\\
    \bottomrule
  \end{tabular}
  \vspace{2mm}
  \caption{\textbf{Necessity of real data.} Our \textit{TrackDiffusion} achieves consistent improvement over various real subsets, especially when there are fewer annotations.}
  \vspace{-5mm}
  \label{tab:necessity}
\end{table}

\section{More Discussion}
\subsubsection{Limitation.}
Limited by training resources, the model's training on a large-scale video dataset was constrained, affecting detail fidelity in video generation. Future work will focus on enhancing detail refinement and broadening generalization to diverse scenarios.

Currently, videos generated by \textit{TrackDiffusion} can only contribute as the augmented samples to train MOT trackers with real data, and it is appealing to explore more flexible usage of the generated video sequences beyond data augmentation, especially the incorporation with generative pre-training~\cite{chen2023mixed,zhili2023task} and contrastive learning~\cite{chen2021multisiam,liu2022task}.
How to generate the high-quality videos aligned with human values without harmful and toxic content~\cite{chen2023gaining,gou2023mixture,gou2024eyes} is also important for the practical usage of \textit{TrackDiffusion}.

\section{More Qualitative Results}
\label{app:qualitative}
\begin{figure}[t]
    \centering
    \includegraphics[width=\linewidth]{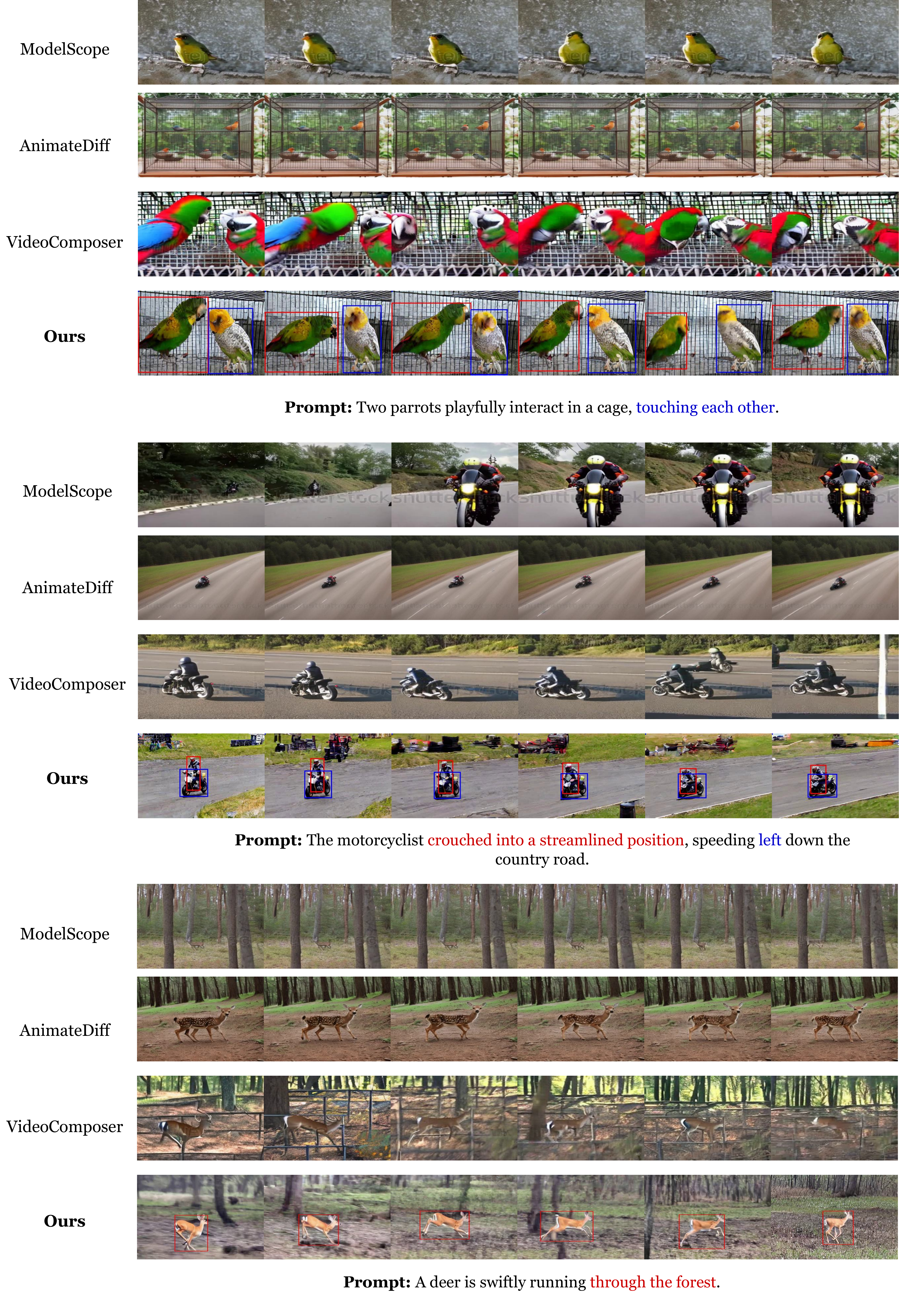}
    \caption{\textbf{Qualitative comparison on the trajectory-conditioned motion control}. 
    ModelScope and AnimateDiff~\cite{guo2023animatediff} do not support controls other than text.}
    \label{fig:app1}
    \vspace{-2mm}
\end{figure}

\begin{figure}[t]
    \centering
    \includegraphics[width=\linewidth]{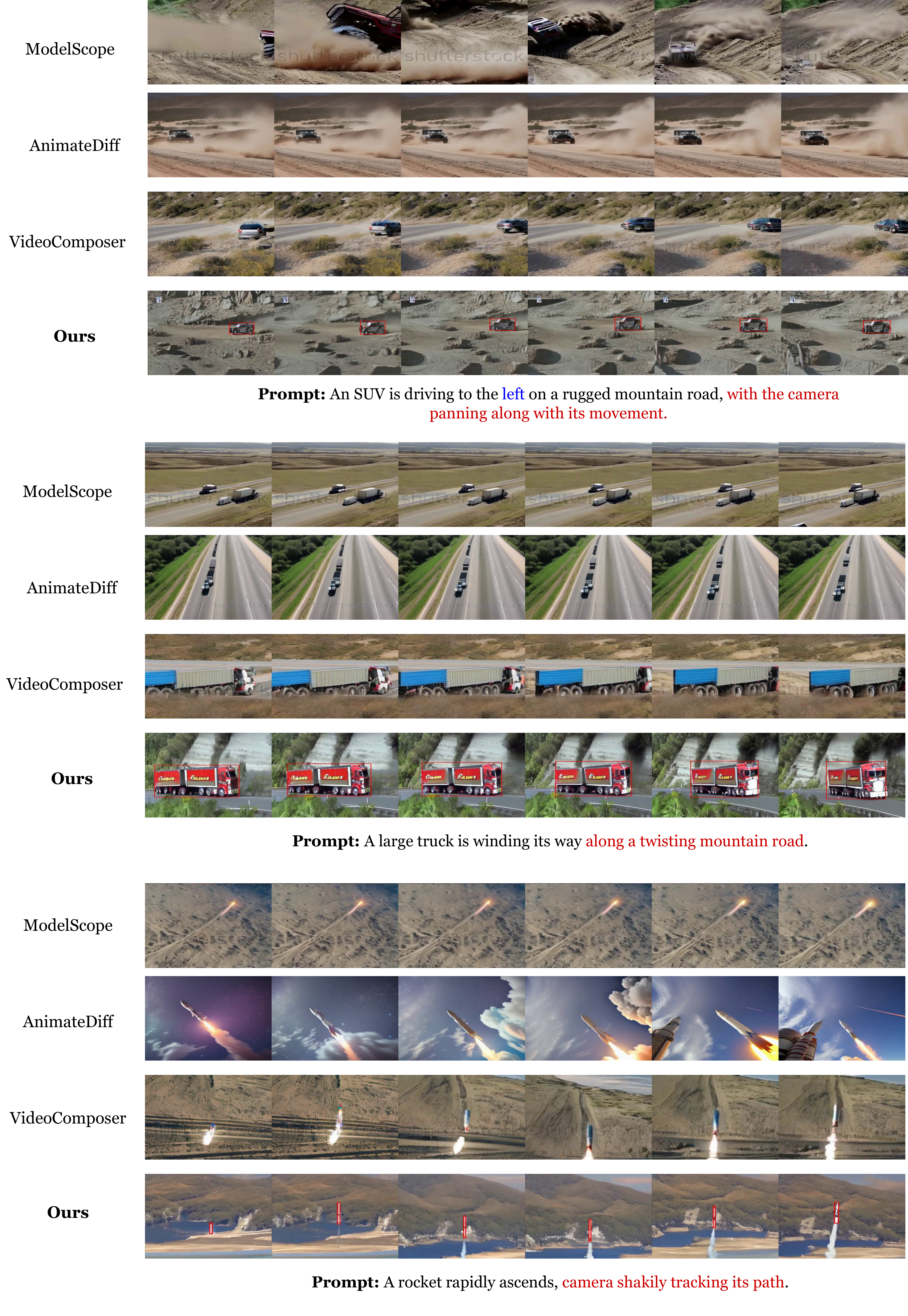}
    \caption{\textbf{Qualitative comparison on the trajectory-conditioned motion control}.}
    \label{fig:app2}
    \vspace{-2mm}
\end{figure}

We provide more qualitative results, which are depicted in Fig.~\ref{fig:app1} and ~\ref{fig:app2}. These figures showcase the effectiveness of our approach in different contexts, highlighting its performance across various scenarios. 

\end{document}